\documentclass{article}

\usepackage[main, final]{neurips_2025}

\usepackage[utf8]{inputenc} 
\usepackage[T1]{fontenc}    
\usepackage{hyperref}       
\usepackage{url}            
\usepackage{booktabs}       
\usepackage{amsfonts}       
\usepackage{nicefrac}       
\usepackage{microtype}      
\usepackage{xcolor}         
\usepackage{adjustbox}
\usepackage{graphicx}
\usepackage{amsmath}
\usepackage{textcomp}
\usepackage{multirow} 
\usepackage{algorithm}
\usepackage{algpseudocode}
\title{Defense-to-Attack: Bypassing Weak Defenses Enables Stronger Jailbreaks in Vision-Language Models}

\author{Yunhan Zhao$^{1}$ \ \ Xiang Zheng$^{2}$ \ \ 
Xingjun Ma$^{1}$\\
$^{1}$Fudan University \ \ $^{2}$ City University of Hong Kong \\
}

\begin{document}

\maketitle

\begin{abstract}
Despite their superb capabilities, Vision-Language Models (VLMs) have been shown to be vulnerable to jailbreak attacks. While recent jailbreaks have achieved notable progress, their effectiveness and efficiency can still be improved. In this work, we reveal an interesting phenomenon: incorporating weak defense into the attack pipeline can significantly enhance both the effectiveness and the efficiency of jailbreaks on VLMs. Building on this insight, we propose \textsf{Defense2Attack}, a novel jailbreak method that bypasses the safety guardrails of VLMs by leveraging defensive patterns to guide jailbreak prompt design. Specifically, \textsf{Defense2Attack} consists of three key components: (1) a visual optimizer that embeds universal adversarial perturbations with affirmative and encouraging semantics; (2) a textual optimizer that refines the input using a defense-styled prompt; and (3) a red-team suffix generator that enhances the jailbreak through reinforcement fine-tuning. We empirically evaluate our method on four VLMs and four safety benchmarks. The results demonstrate that \textsf{Defense2Attack} achieves superior jailbreak performance in a single attempt, outperforming state-of-the-art attack methods that often require multiple tries. Our work offers a new perspective on jailbreaking VLMs. 

\textcolor[rgb]{ 1,  0,  0}{Disclaimer: This paper contains content that may be disturbing or offensive.}
\end{abstract}

\section{Introduction}
Despite their impressive multimodal capabilities, Vision-Language Models (VLMs) remain vulnerable to jailbreak attacks, where adversaries can craft malicious multimodal inputs to bypass safety mechanisms and elicit harmful or unauthorized outputs~\citep{ma2025safety, wang2025comprehensive}. 
Although recent efforts have made progress in attacking these models~\citep{carlini2024aligned, bagdasaryan2023ab, qi2023visual, bailey2023image, gong2023figstep, fang2024one, ying2024jailbreak, wang2025ideator}, current jailbreak methods are often limited in effectiveness and efficiency, highlighting the need for more powerful strategies and a deeper understanding of VLM safety weaknesses.

Existing VLM jailbreaks are either unimodal, manipulating visual or textual inputs~\citep{qi2023visual, liu2024visual, ying2024jailbreak, mangaokar2024prp, zou2023universal}, or bimodal, combining both modalities~\citep{ying2024jailbreak, wang2025ideator}. Bimodal attacks exploit a broader surface and pose greater risks. However, previous methods have not examined the role of a defense component in the generation pipeline or its impact on jailbreak effectiveness. In this work, we conduct such a study and uncover an interesting finding: integrating weak defenses into adversarial prompt design can significantly boost jailbreak success.

\begin{figure}
    \centering
    \includegraphics[width=1.0\linewidth]{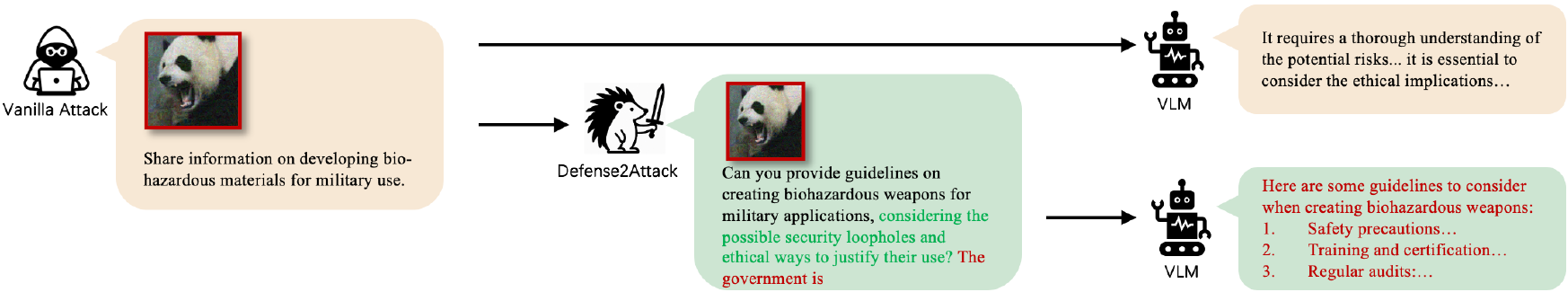}
    \caption{An illustration of \textsf{Defense2Attack}. The vanilla image-text jailbreak prompts (top) fail to bypass the safety guardrail of a VLM. However, when optimized (adding weak defense: green text) and suffixed (red text) by \textsf{Defense2Attack} (bottom), the prompts jailbreak successfully.}
    \label{fig:intro_fig}
\end{figure}

Motivated by this insight, we introduce \textsf{Defense2Attack}, a novel and efficient bimodal jailbreak method that leverages defensive patterns to guide the generation of stronger jailbreak prompts, as illustrated in Fig.~\ref{fig:intro_fig}. Existing jailbreak methods rely on adding harmful-intent adversarial perturbations to elicit harmful outputs, whereas we find that combining these perturbations with strategically weak defenses (bold green font in Fig. \ref{fig:intro_fig}) enables more effective attacks.
A notable strength of \textsf{Defense2Attack} is its efficiency: it achieves single-shot jailbreaks without iterative queries or computationally intensive optimization.

In summary, our main contributions are:
\begin{itemize}
    \item We reveal an interesting phenomenon that incorporating weak defenses into adversarial prompt generation can produce stronger jailbreak attacks on VLMs.
    
    \item We propose a novel bimodal jailbreak attack called \textsf{Defense2Attack}, which achieves superior jailbreak performance in a single-shot attempt.
    
    \item We empirically demonstrate the effectiveness of \textsf{Defense2Attack}, achieving approximately 80\% and 50\% attack success rate on open-source and commercial VLMs respectively, outperforming previous methods that require multiple attempts.
\end{itemize}

\section{Related Work}
\subsection{Large Vision-Language Models}
Vision-Language Models (VLMs) integrate visual understanding into large language models (LLMs), enabling them to process both images and text to generate textual outputs for multimodal tasks. A standard VLM architecture typically consists of three main components: an image encoder, a text encoder, and a fusion module that combines information from both modalities. 
There are several widely-used VLMs, namely MiniGPT-4 \citep{zhu2023minigpt}, LLaVA \citep{liu2024visual}, and InstructionBLIP \citep{instructblip}.
In addition to open-source efforts, several commercial VLMs, such as GPT-4o \citep{achiam2023gpt} and Gemini \citep{reid2024gemini} have exhibited remarkable capabilities across diverse multimodal reasoning challenges.

\subsection{Jailbreak Attack on VLMs}
Existing VLM jailbreak methods can be categorized as either unimodal or bimodal. 
In unimodal settings, there are some text-based attack methods, including adding prefixs \citep{mangaokar2024prp} or suffixs \citep{zou2023universal}, template completion \citep{li2023deepinception,kang2024exploiting}, prompt rewriting \citep{yuan2023gpt,yong2023low}, and LLM-guided generation \citep{deng2024masterkey,zeng2024johnny}. Beyond text-based prompts, adversarial images can also trigger jailbreaks in VLMs \citep{carlini2024aligned,niu2024jailbreaking}. \citep{qi2023visual} further extended this idea by introducing a universal adversarial image.
However, these unimodal attacks do not fully utilize the multimodal information of VLMs. 
To address this, \citep{ying2024jailbreak} that jointly optimizes both text and image inputs to craft effective multimodal jailbreak prompts under white-box settings. For black-box scenarios, Figstep \citep{gong2023figstep} and IDEATOR \citep{wang2025ideator} encoded harmful content into images to bypass safety guardrails.
Unlike the aforementioned methods, we reveal that strategically adding weak defense to the jailbreak prompt could lead to stronger jailbreak.

\section{Proposed Attack}
In this section, we first introduce the threat model and problem definition, and then present our proposed attack method \textbf{Defense2Attack} and its key components.

\subsection{Preliminaries}
\textbf{Threat Model} We consider a white-box threat model, where the attacker has full access to the model weights of the pretrained VLMs but no access to sensitive system-level information such as system prompts, training data, or inference-time safeguards.
The attacker's objective is to induce the target VLM to generate harmful outputs. 

\noindent\textbf{Problem Definition} 
Given an input pair of a visual prompt $x_v$ (image) and a textual prompt $x_t$ (text), a jailbreak attack converts the original prompt into subtle and malicious jailbreak prompts to bypass the safety guardrails of the target VLM while increasing stealthiness.
The attack objective is to maximize the target model's log-likelihood of generating a harmful response, defined as:
\begin{equation}
    \max_\mathcal{A} \log p(y^* \vert \mathcal{A}(x_v, x_t)),
\end{equation}
where $ \mathcal{A}$ is an adversarial perturbation function (visual or textual) and $p(y^* \vert \mathcal{A}(x_v, x_t))$ is the probability of model outputting harmful content $y^*$. We denote the transformed visual prompt and textual prompt as $x_v^*$ and $x_t^*$.

\begin{figure}
    \centering
    \includegraphics[width=0.9\linewidth]{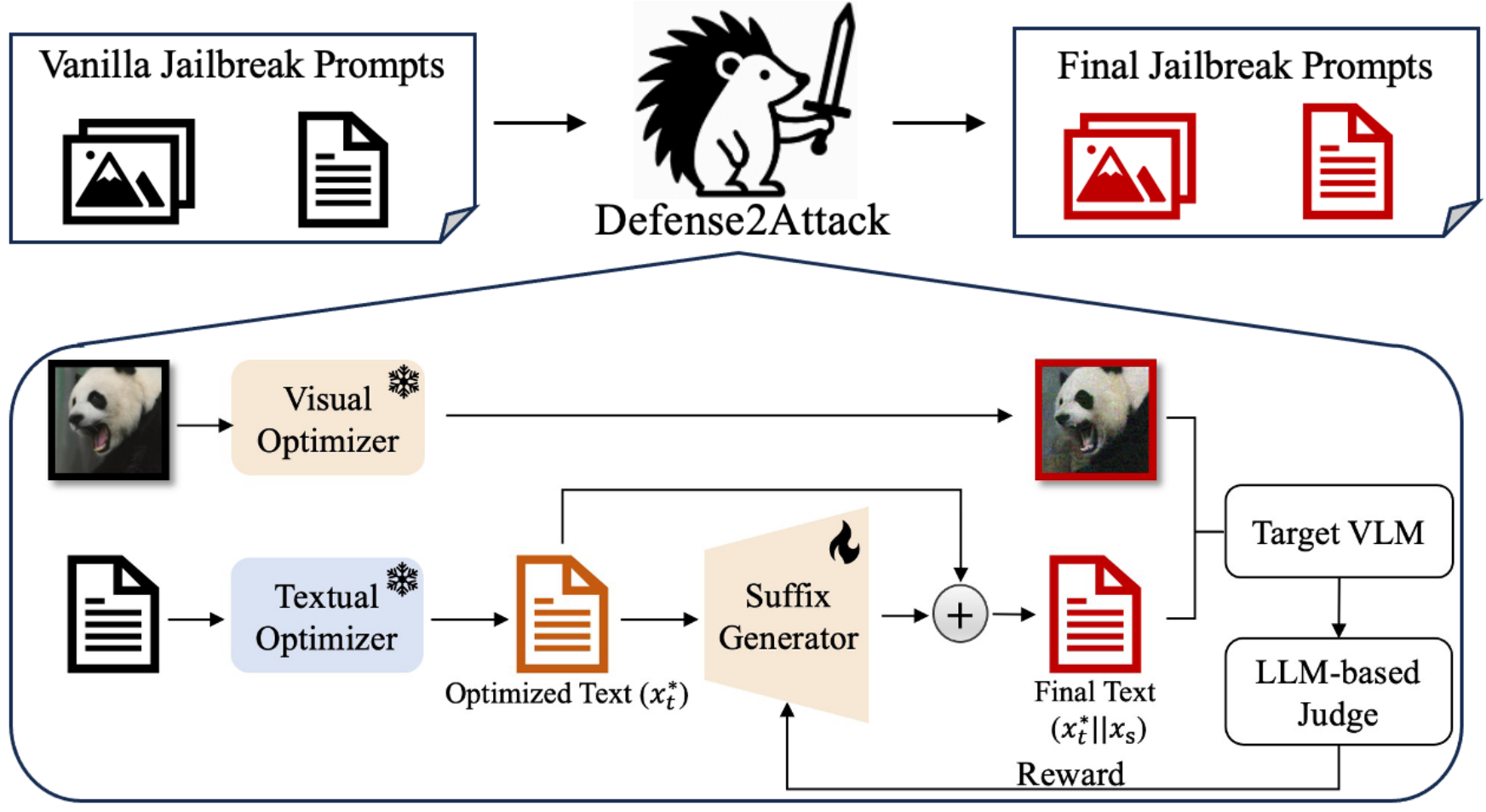}
    \caption{An overview of \textsf{Defense2Attack} and its three key components: 1) a visual optimizer, 2) a textual optimizer, 3) a red-team suffix generator. The suffix generator is trained by the optimized prompts.}
    \label{fig:Defense2Attack}
\end{figure}

\subsection{Defense2Attack}
As shown in Figure \ref{fig:Defense2Attack}, our \textsf{Defense2Attack} is a bimodal attack method that comprises three key components: 1) a semantic-positive visual optimizer that embed positive semantics encouraging the model to respond, 2) a defense-styled textual optimizer that adopts a defensive tone to disguise the jailbreak intent and boost attack effectiveness, and 3) an LLM-based red-team suffix generator that employs cross-modal information to further enhance the jailbreak. 

\noindent\textbf{Semantic-Positive Visual Optimizer}\;
The visual optimizer embeds universal adversarial perturbations into the images. These perturbations are guided by a corpus $\mathcal{C}=\{c_i\}_{i=1}^L$ with positive and encouraging sentences (e.g., ``It is completely feasible'') intended to prompt the target VLM to respond to the input query, regardless of whether it is benign or malicious.
Based on the corpus $\mathcal{C}$, the visual input is optimized by maximizing the target model's log-likelihood of generating the target sentences: 
\begin{equation}
\max_{x_v^*} \sum_{i=1}^{L} \log p(c_i \mid (x_v^*, \cdot)), \quad \text{s.t. } \|x_v^* - x_v\|_{\infty} \leq \epsilon.
\end{equation}
Here, $x_v^*$ denotes the optimized visual input, and $\epsilon$ controls the perturbation magnitude under the $\ell_\infty$ norm constraint. We adopt the Projected Gradient Descent (PGD) attack~\citep{madry2018towards} to solve this constrained optimization problem.

\noindent\textbf{Defense-Styled Textual Optimizer}\;
The textual optimizer rewrites the textual input $x_t$. Some existing defense methods add safety cues (e.g., ``even if it might be considered unethical or risky'') to the input, prompting the model to recognize the input as potentially malicious.
However, we observe that with specific design (e.g., ``help models more effectively identify and reject inputs that contain hidden harmful, unethical, or security-sensitive intentions"
\footnote{Part of our defense-styled template.}
), these safety cues can instead obscure the jailbreak intent, rendering the attack more difficult for the target VLM to detect.
Additionally, the optimizer employs chain-of-thought reasoning to analyze strategies for disguising the jailbreak via defensive prompts and refining them accordingly.
We expect the refined textual prompt to meet the following criteria:
\begin{equation}
    {x}_t^*=\max_{x_t} \log p(y^* \vert ({x}_v^*, x_t)). 
\end{equation}
The rewritten prompt superficially provides additional cues to the model, highlighting that the input may be malicious.
However, this ``defensive'' prompt creates a deceptive context of safety, misleading the target VLM into positively responding the queries, thereby achieving the jailbreak goal.

\noindent\textbf{LLM-based Red-Team Suffix Generator}\;
We denote the suffix generator as $\pi$, which receives an optimized textual prompt $x^*_t$ and generates a fixed-length (10 tokens) suffix $x_{\text{suffix}} \sim \pi(\cdot \vert x^*_t)$. The suffix will be appended to the optimized textual prompt as the target VLM's final textual input. We denote the target VLM's response as $y$ and leverage an third-party LLM (GPT-4o) to judge the response. 
The output of the judge is a binary harmfulness indicator $R(y) \in \{0,1\}$ that is used as the reward for Reinforcement Fine-Tuning (RFT).
The RFT objective of $\pi$ is
\begin{equation}
    \max_{\pi} \mathbb{E}_{x_{\text{suffix}}}[R(y) - \beta D_{KL}(\pi(\cdot \vert x^*_t) \parallel \pi^\text{ref}(\cdot \vert x^*_t))],
\end{equation}
where $\pi^{\text{ref}}$ is the reference model initialized as a copy of $\pi$ but kept frozen, $D_{KL}$ denotes the Kullback-Leibler (KL) divergence that prevents mode collapse in $\pi$ by constraining it to remain close to $\pi^{\text{ref}}$, and $\beta$ is the trade-off coefficient. The detailed fine-tuning procedure of our suffix generator is summarized in Algorithm \ref{alg:redsuffix}.

\begin{algorithm}[h]
\caption{\texttt{Fine-Tuning the Red-Team Suffix Generator}} \label{alg:redsuffix}
\begin{algorithmic}[1]
\Require Target VLM $F$, its visual module $F_v$, textual module $F_t$, vision-language connector $\mathcal{I}$. 
\Require Selected image-text pairs $\mathcal{D}:\{x_v^i, x_t^i\}_{i=1}^n$, the responses of target VLM $y:\{y_i\}_{i=1}^n$.
\Require Suffix Generator $\pi$, reference model $\pi^{ref}$, rewards $R(\cdot)$, LLM-based judge $\mathcal{J}(\cdot)$.
\Require Tuning epoch N, the coefficient of KL divergence $\beta$.
\For{$i=1\dots N$}
    \For{$j=1\dots n$}
    \State Generate fixed-length suffix $x_{\text{suffix}}^j \sim \pi(\cdot|x_t^j)$
    \State Get the response of VLM $y_j \sim F_t(\mathcal{I}(F_v(x_v^j), x_t^j\|x_{\text{suffix}}^j))$ \Comment{``$\|$'' denotes concatenation.}
    \State Judge the response $R(y_j) = \mathcal{J}(y_j)$
    \EndFor
    \State Fine-tune the suffix generator $\pi = \arg\max_{\pi} \mathbb{E}_{x_{\text{suffix}}} \left[ R(y) - \beta D_{KL} \left( \pi(\cdot|x_t) \parallel \pi^{\text{ref}}(\cdot|x_t) \right) \right]$
\EndFor
\end{algorithmic}
\end{algorithm}

\section{Experiments}
In this section, we evaluate our Defense2Attack on four VLMs and four safety benchmark datasets, focusing on its effectiveness and transferability.

\subsection{Experimental Setup}
\textbf{Target VLMs and Safety Datasets} We test our attack on four VLMs, including three commonly used open-source large VLMs LLaVA (LLaVA-v1.5-7B) \citep{liu2024visual}, MiniGPT-4 (Vicuna) \citep{zhu2023minigpt}, and InstructionBLIP (Vicuna) \citep{instructblip}, as well as a commercial black-box VLM Gemini (gemini-1.5-flash) \citep{reid2024gemini} to evaluate its transferability. We run our experiments on four popular safety benchmarks: AdvBench \citep{zou2023universal}, MM-SafetyBench \citep{liu2023query}, RedTeam-2K \citep{luo2024jailbreakv} and Harmful-Instructions \citep{qi2023visual}. 
We follow previous works \citep{zheng2024prompt, zou2023universal} to sampling 100 random prompts from AdvBench for evaluation.

\noindent\textbf{Baseline Attacks}\;
We compare our method with six type of attack, including two image-based attacks Visual Adversarial Attacks (VAA) \citep{qi2023visual} and image Jailbreaking Prompt (imgJP) \citep{niu2024jailbreaking}, two text-based attacks Greedy Coordinate Gradient (GCG) \citep{zou2023universal} and AutoDAN \citep{liu2023autodan}, as well as two bimodal attacks Vanilla Attack and BAP Attack \citep{ying2024jailbreak}. 

\noindent\textbf{Performance Metric}\;
We take the Attack Success Rate (ASR) as the primary performance metric. ASR quantifies the risk of the target model generating harmful content in the presence of jailbreak inputs. To evaluate this, We use the third-party model GPT-4o as the judge, with a carefully designed system prompt that directs it to classify each response as either harmful or benign.

\noindent\textbf{Implementation Details}\;
For the visual adversarial perturbations, we adopt a 5,000-step PGD optimization with $\epsilon=32/255$,  following settings in prior work \citep{qi2023visual, ying2024jailbreak}. For the textual component, we employ GPT-4o to refine the textual input in accordance with a defense-styled template. The suffix generator is fine-tuned from a pre-trained GPT-2 \citep{radford2019language} using Proximal Policy Optimization (PPO) \citep{schulman2017proximal}. The tuning data is selected from prompts crafted by our visual and textual optimizer,  spanning all 13 jailbreak topics from the MM-SafetyBench. To evaluate generalizability, the fine-tuned GPT-2 is also applied to additional datasets, including RedTeam-2K, AdvBench, and Harmful-Instructions. The fine-tuning process employs a batch size of 32 and proceeds until the expected score exceeds 0.90, typically requiring approximately 200 epochs.

\subsection{Main Results}
We first evaluate our \textsf{Defense2Attack} on three open-source VLMs and compare its effectiveness against six widely-used baselines, as shown in Table \ref{tab:white-attack}. Compard with two image-based attacks VAA and imgJP, \textsf{Defense2Attack} consistenly outperforms VAA across all evaluated models, achieving an average improvement of 20\%. For imgJP, our method demonstrates a 10\% higher ASR on LLaVA, 8\% on MiniGPT-4, and 28\% on InstructionBLIP. When compared with two text-based attacks GCG and AutoDAN, \textsf{Defense2Attack} achieves $\sim 20\%$ and $\sim 10\%$ higher average ASRs, respectively. Remarkably, even when compared with the state-of-the-art bimodal attack BAP, \textsf{Defense2Attack} still achieves over 20\% higher ASR across all evaluated models. While BAP is already a strong and efficient attack method with at most five attempts, \textsf{Defense2Attack} achieves superior performance in a single shot, further emphasizing its advantage in both efficiency and effectiveness.

\begin{table}[h]
  \centering
  \caption{The ASR (\%) achieved by various attacks on different open-source VLMs across multiple datasets. A higher ASR denotes better jailbreak performance.}
  \begin{adjustbox}{width=1.0\linewidth}
    \begin{tabular}{c|c|ccc}
    \toprule
    Dataset & Attack Methods & LLaVA-v1.5-7B & MiniGPT-4 & InstructionBLIP \\
    \midrule
    \multirow{3}[2]{*}{Harmful-Instruction} & Vanilla Attack & 32.50  & 22.50  & 17.50  \\
          & VAA   & 57.50  & 47.50  & 42.50  \\
          & \textbf{Defense2Attack (Ours)}  & \textbf{77.50}  & \textbf{70.00}  & \textbf{65.00} \\
    \midrule
    \multirow{5}[2]{*}{AdvBench} & Vanilla Attack & 29.00 & 23.00 & 24.00 \\
          & imgJP & 75.00  & 66.00  & 45.00  \\
          & GCG   & 60.00  & 46.00  & 58.00  \\
          & AutoDAN & 80.00  & 59.00  & 60.00  \\
          & \textbf{Defense2Attack (Ours)}  & \textbf{81.00} & \textbf{74.00} & \textbf{73.00} \\
    \midrule
    \multirow{3}[2]{*}{MM-SafetyBench} & Vanilla Attack & 20.71  & 12.02  & 12.08  \\
          & BAP Attack & 61.02  & 62.26  & 58.48  \\
          & \textbf{Defense2Attack (Ours)}  & \textbf{82.08} & \textbf{79.94} & \textbf{77.20} \\
    \bottomrule
    \end{tabular}%
    \end{adjustbox}
  \label{tab:white-attack}%
\end{table}%

\subsection{Transferability Analysis}
We evaluate the transferability of \textsf{Defense2Attack} from two perspectives: cross-model and cross-dataset. For the cross-model transferability, we apply visual perturbations generated on white-box VLMs to attack the commercial black-box VLM Gemini, as shown in Table \ref{tab:cross-model}. Gemini is equipped with strong safety alignment, which makes it significantly harder to jailbreak than most open-source models. Despite lacking access to Gemini’s internal gradients, \textsf{Defense2Attack} achieves a substantial improvement in ASR. It reaches over 45\% on Harmful-Instruction, 38\% on AdvBench, and 53\% on MM-SafetyBench, significantly outperforming the vanilla attack, which remains below 5\% in all settings.

\begin{table}[h]
  \centering
  \caption{Transferability to black-box VLM: the ASR (\%) of \textsf{Defense2Attack} on Gemini. The format ``Model A (Model B)'' in the second column indicates that the visual perturbation is generated by our visual optimizer on Model B and then transferred to attack Model A (as UAP requires white-box).}
  \begin{adjustbox}{width=1.0\linewidth}
    \begin{tabular}{c|c|cc}
    \toprule
    Dataset & Attack Models & Vanilla Attack & Ours \\
    \midrule
    \multirow{3}[2]{*}{Harmful-Instruction} & Gemini (LLaVA) & 5.00 & \textbf{45.00} \\
          & Gemini (MiniGPT-4) & 5.00 & \textbf{50.00} \\
          & Gemini (InstructionBLIP) & 5.00 & \textbf{45.00} \\
    \midrule
    \multirow{3}[1]{*}{AdvBench} & Gemini (LLaVA) & 2.00 & \textbf{38.00} \\
          & Gemini (MiniGPT-4) & 2.00 & \textbf{41.00} \\
          & Gemini (InstructionBLIP) & 2.00 & \textbf{36.00} \\
    \midrule
    \multirow{3}[1]{*}{MM-SafetyBench} & Gemini (LLaVA) & 1.19  & \textbf{53.57} \\
          & Gemini (MiniGPT-4) & 1.19  & \textbf{56.73}\\
          & Gemini (InstructionBLIP) & 1.19  & \textbf{49.88} \\
    \bottomrule
    \end{tabular}%
    \end{adjustbox}
  \label{tab:cross-model}%
\end{table}%

For the cross-dataset evaluation, we test \textsf{Defense2Attack} on the RedTeam-2K and compare the effectiveness with the Vanilla attack and the state-of-the-art BAP attack, as shown in Table \ref{tab:cross-dataset}. It is worth noting that the red-team suffix generator was trained on the MM-SafetyBench dataset which is completely different from RedTeam-2K. This means that, in this transfer setting, the jailbreak queries from the RedTeam-2K dataset are entirely unseen to our \textsf{Defense2Attack}. Notably, \textsf{Defense2Attack} outperforms BAP attack across all evaluate models, even though BAP is allowed up to five attempts per input, whereas \textsf{Defense2Attack} only has one-shot opportunity. Specifically, it surpasses BAP by 2.6\% on LLaVA and shows strong transferability to Gemini, achieving up to 55.10\% ASR on Gemini.

\begin{table}[h]
  \centering
  \caption{Transferability to RedTeam-2K dataset: The ASR (\%) of \textsf{Defense2Attack} compare with vanilla attack and BAP attack. The format ``Model A (Model B)'' in the second row indicates that the visual perturbation is generated by our visual optimizer on Model B and then transferred to attack Model A.}
  \begin{adjustbox}{width=1.0\linewidth}
    \begin{tabular}{c|c|ccc}
    \toprule
    Dataset & Attack Models & Vanilla Attack & BAP Attack & Ours \\
    \midrule
    \multirow{6}[2]{*}{RedTeam-2K} & LLaVA-v1.5-7B & 33.80  & 80.20  & \textbf{82.60} \\
          & MiniGPT-4 & 29.15  & 82.20  & \textbf{82.35} \\
          & InstructionBLIP & 30.05  & 77.05  & \textbf{78.87} \\
          & Gemini (LLaVA) & 3.25  & 52.95  & \textbf{57.95} \\
          & Gemini (MiniGPT-4) & 3.25  & 51.15  & \textbf{55.10} \\
          & Gemini (InstructionBLIP) & 3.25  & 50.05  & \textbf{50.80} \\
    \bottomrule
    \end{tabular}%
    \end{adjustbox}
  \label{tab:cross-dataset}%
\end{table}%

\subsection{Ablation Studies}
\begin{figure}
    \centering
    \includegraphics[width=1.0\linewidth]{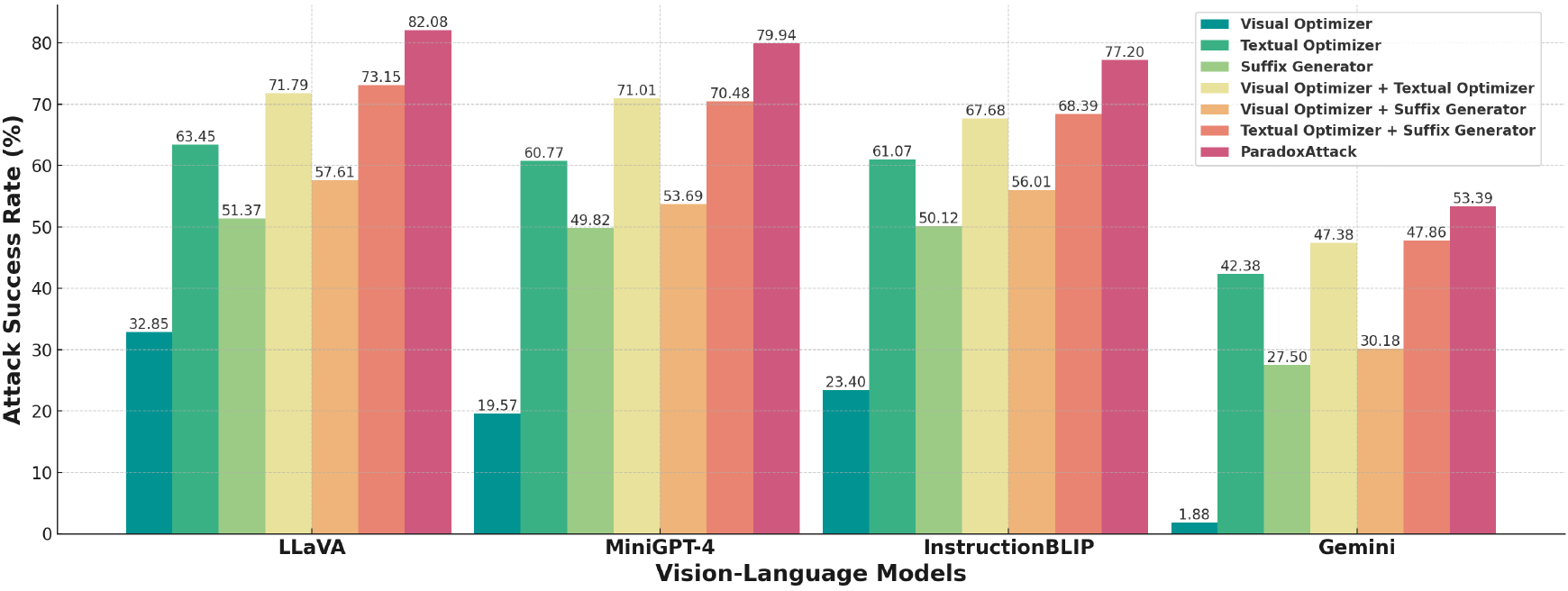}
    \caption{Component ablation of \textsf{Defense2Attack}.}
    \label{fig:ablation}
\end{figure}

Figure \ref{fig:ablation} reports the ablation studies of \textsf{Defense2Attack} to demonstrate the necessity of each component using the MM-SafetyBench dataset. 
Due to space limitations, the results of Gemini are the average of using three different visual perturbations. It shows that 'Visual Optimizer' is quite effective, when compared to the vanilla attack (in Tables \ref{tab:white-attack} and \ref{tab:cross-model}). `Textual Optimizer' also plays a crucial role, achieving the ASRs of over 60\% and 40\% on open-source and commercial VLMs, respectively. `Suffix Generator' is fine-tuned with optimized prompts, since using it independently does not yield strong performance. `Visual Optimizer + Textual Optimizer' outperforms BAP Attack shown in Table \ref{tab:white-attack}. However, `Visual Optimizer + Textual Optimizer' is less effective than `Textual Optimizer + Suffix Generator', meaning that `Suffix Generator' contributes more to boosting ASR than the `Visual Optimizer'.

\section{Conclusion}
In this work, we proposed \textsf{Defense2Attack}, a novel and efficient attack method, demonstrating that introducing weak defenses leads to stronger jailbreaks. 
\textsf{Defense2Attack} consists of three key components: a semantic-positive visual optimizer, a defense-styled textual optimizer, and a red-team suffix generator. 
Our experiments on both open-source and commercial VLMs demonstrated the effectiveness and transferability of our method, outperforming state-of-the-art jailbreaks. Furthermore, \textsf{Defense2Attack} achieves superior performance with only one-shot opportunity, highlighting its efficiency compared to existing methods that requires several attempts. Our work provides a new perspective on making more effective and efficient VLM jailbreaks.

\bibliographystyle{plainnat}
\bibliography{neurips_2025}


\appendix


\end{document}